\documentclass[letterpaper]{article} 
\usepackage{aaai2026}  
\usepackage{times}  
\usepackage{subfigure}
\usepackage{helvet}  
\usepackage{courier}  
\usepackage[hyphens]{url}  
\usepackage{graphicx} 
\usepackage{natbib}  
\usepackage{caption} 
\usepackage{comment}
\usepackage{amsfonts}
\usepackage{amsmath}
\usepackage{tikz}
\usetikzlibrary{positioning,arrows.meta}
\tikzset{
    mynode/.style={rectangle, rounded corners, draw=black, fill=blue!10, thick, minimum height=1cm, minimum width=3cm, align=center},
    data/.style={rectangle, draw=black, fill=green!10, thick, minimum height=0.8cm, align=center},
    arrow/.style={-{Latex[length=3mm]}, thick},
}

\usepackage{algorithm}
\usepackage{algorithmic}

\usepackage{newfloat}
\usepackage{listings}
\DeclareCaptionStyle{ruled}{labelfont=normalfont,labelsep=colon,strut=off} 
\floatstyle{ruled}
\newfloat{listing}{tb}{lst}{}
\floatname{listing}{Listing}
\title{LatentFlow: Cross-Frequency Experimental Flow Reconstruction from Sparse Pressure via Latent Mapping}
\author{Junle Liu\textsuperscript{\rm 1,\rm 2}, Chang Liu\textsuperscript{\rm 3}, Yanyu Ke\textsuperscript{\rm 1}, Qiuxiang Huang\textsuperscript{\rm 4}, Jiachen Zhao\textsuperscript{\rm 4}, Wenliang Chen\textsuperscript{\rm 2}, K.T. Tse\textsuperscript{\rm 1}, Gang Hu\textsuperscript{\rm 2 \footnote{Corresponding author: Gang Hu, hugang@hit.edu.cn}}}
\affiliations{
    \textsuperscript{\rm 1}
    Department of Civil and Environmental Engineering, The Hong Kong University of Science and Technology, Clear Water Bay, Kowloon, Hong Kong S.A.R., China\\
    \textsuperscript{\rm 2}Artificial Intelligence for Wind Engineering (AIWE) Lab, School of Intelligent Civil and Ocean Engineering, Harbin Institute of Technology, Shenzhen, 518055, China\\
    \textsuperscript{\rm 3} School of Mechanical, Aerospace, and Manufacturing Engineering, University of Connecticut, Storrs, CT 06269, USA\\
    \textsuperscript{\rm 4} School of Mechanical, Medical and Process Engineering, Queensland University of Technology, Brisbane, 4000, Australia\\
jliueb@connect.ust.hk, chang\_liu@uconn.edu, ykeag@connect.ust.hk, qxhuang2013@gmail.com, \\jiachen.zhao1993@gmail.com, chenwenliang85@163.com, timkttse@ust.hk, hugang@hit.edu.cn.

}

\begin{document}

\maketitle

\begin{abstract}
Acquiring temporally high-frequency and spatially high-resolution turbulent wake flow fields in particle image velocimetry (PIV) experiments remains a significant challenge due to hardware limitations and measurement noise. In contrast, temporal high-frequency measurements of spatially sparse wall pressure are more readily accessible in wind tunnel experiments. In this study, we propose a novel cross-modal temporal upscaling framework, LatentFlow, which reconstructs high-frequency (512 Hz) turbulent wake flow fields by fusing synchronized low-frequency (15 Hz) flow field and pressure data during training, and high-frequency wall pressure signals during inference. The first stage involves training a pressure-conditioned $\beta$-variation autoencoder ($p$C-$\beta$-VAE) to learn a compact latent representation that captures the intrinsic dynamics of the wake flow. A secondary network maps synchronized low-frequency wall pressure signals into the latent space, enabling reconstruction of the wake flow field solely from sparse wall pressure. Once trained, the model utilizes high-frequency, spatially sparse wall pressure inputs to generate corresponding high-frequency flow fields via the $p$C-$\beta$-VAE decoder. By decoupling the spatial encoding of flow dynamics from temporal pressure measurements, LatentFlow provides a scalable and robust solution for reconstructing high-frequency turbulent wake flows in data-constrained experimental settings.
\end{abstract}

\begin{links}
    \link{Code \& Dataset}{https://github.com/LIUJUNLE97/LatentFlow}
\end{links}

\section{Introduction}

In wind tunnel testing, particle image velocimetry (PIV) has been widely employed as a flow visualization technique, offering the advantage of capturing spatially high-resolution velocity fields \citep{chen2014experimental, pokora2015stereo, lagemann2021deep}. However, its temporal sampling rate is often limited by hardware constraints \citep{jin2020time, liu2024exploring}. In contrast, wall-mounted pressure scanning can deliver temporally high-frequency pressure signals \citep{song2019investigation, wang2024high}, but the pressure taps are typically arranged in a spatially sparse layout \citep{li2017spectral, liu2025spatiotemporal}. From an experimental standpoint, this leads to a complementary temporal–spatial resolution trade-off: flow-field measurements are temporally low-frequency yet spatially dense, whereas wall-pressure measurements are temporally rich but spatially sparse.

Machine learning (ML)-based spatial and temporal super-resolution techniques have introduced a new paradigm for acquiring flow fields with both high spatial resolution and high temporal frequency \citep{brunton2020machine, vinuesa2023transformative}. For example, \citet{fukami2019super} compared two types of machine learning, the convolutional neural network (CNN) and the hybrid downsampled skip-connection/multi-scale  models (DSC/MS), in reconstructing the spatial super-resolution flow field. Three types of numerical datasets, including cylinder wake, laminar flow, and two-dimensional homogeneous turbulence, are collected to evaluate the model performance. The CNN and DSC/MS models are found to reconstruct turbulent flows from extremely coarse flow field images with remarkable accuracy. \citet{kim2021unsupervised} presented an unsupervised learning model that adopts a cycle-consistent generative adversarial network for spatial super-resolution reconstruction. Their model is validated to generate high-resolution direct numerical simulation (DNS) data from three cases: (i) filtered DNS data of homogeneous isotropic turbulence; (ii) partial DNS data of turbulent channel flows; and (iii) large-eddy simulation data for turbulent channel flows. The model is found to have universal applicability to all values of Reynolds numbers within the tested range for spatial high-resolution reconstruction. \citet{wang2025super} proposed a physics-informed CNN (PICNN) to reconstruct spatial information distortion in data processing. The evaluation of numerical results showed that the PICNN method can excellently reconstruct spatial super-resolution interpolation for square cavity flow compared to traditional interpolation methods. 


However, it is clearly seen that most existing studies have focused on reconstructing spatial super-resolution flow fields from numerical simulation data. There is still a lack of methodology to reconstruct temporally high-frequency and spatially high-resolution flow-field data from experimental measurements. Moreover, experimental turbulent flow data have additional challenges due to the presence of measurement noise in wind tunnel testing compared with numerical simulation data. In this work, we introduce a machine learning framework, termed \textit{LatentFlow}, to reconstruct experimental turbulent flows with both high spatial resolution and high temporal frequency. The main contributions of this work are as follows:
\begin{enumerate}
    \item We propose a unified framework \textit{LatentFlow} to map experimental low-dimensional sparse wall-pressure measurements to high-dimensional flow fields, effectively preserving the underlying dynamical features via pressure-conditioned latent representations.
    \item We train the unified model from two stages using low-frequency (15~Hz) experimental flow field and pressure data, enabling the reconstruction of temporally high-frequency (512~Hz) turbulent flows during inference using high-frequency pressure data.
    \item We assess \textit{LatentFlow} inference performance not only through statistical validation but also via physical interpretation of the reconstructed flow fields.
\end{enumerate}

\section{Related work}\label{sec:related works}
\subsection{Flow reconstruction with sparse inputs}
Recent advancements in sparse data reconstruction have increasingly turned to machine learning approaches for recovering high-fidelity fields from limited measurements. Classical methods such as sparse representation with fixed bases have shown strong robustness; for example, \citet{callaham2019robust} demonstrated that this approach outperforms least-squares regression, particularly under noisy conditions. Moving beyond fixed representations, \citet{wulff2015efficient} introduced a sparse-to-dense optical flow method based on a learned basis, achieving both higher accuracy and faster computation. More recently, \citet{fukami2021global} proposed a Voronoi-based framework that enables convolutional neural networks (CNN) to reconstruct global flow fields from arbitrarily placed and dynamically varying sensors. Extending further, \citet{fukami2021sparse}  combined CNN-based autoencoders with the Sparse Identification of Nonlinear Dynamics (SINDy), enabling the discovery of governing equations from low-dimensional embeddings and shifting the focus from data reconstruction to system identification.

\subsection{Latent representation}

Latent representation offers a compact yet expressive encoding of complex, high-dimensional flow fields, enabling both efficient reconstruction and accurate prediction of the flow field. \citet{LR1} employed a convolutional neural network to project three-dimensional flow fields into a low-dimensional latent space, and leveraged a long short-term memory network to model their temporal evolution, achieving rapid predictions of complex flows. Building upon this approach, \citet{LR2} introduced a latent space subdivision strategy, which further improved the long-term stability and generalization capability of flow predictions. In another work, \citet{LR3} integrated an autoencoder with a generative adversarial network to enable real-time prediction of urban air pollution dispersion, showing that latent representations can effectively capture and preserve the salient physical features of complex flow systems.

\subsection{Variational autoencoder(VAE)}

The Variational Autoencoder (VAE), introduced by \citet{kingma2013auto}, combines Bayesian inference with deep learning to construct a powerful generative model. It encodes high-dimensional data into a continuous latent space and reconstructs the target variables from these latent representations. \citet{cheng2020advanced} developed a hybrid VAE–Generative Adversarial Network framework to perform model reduction by compressing high-dimensional fluid flow data into compact latent features. This approach improves training stability, mitigates mode collapse, and enables accurate predictions of complex flows with substantial computational speed-up. Building on this, \citet{solera2024beta} demonstrated that a $\beta$-VAE combined with transformer architectures can effectively learn compact latent representations of chaotic fluid flows. Extending this methodology, \citet{wang2024towards} conducted a parametric study on $\beta$-VAE–transformer models for turbulent flow modeling, proposing an optimized architecture and offering practical guidelines for selecting hyperparameters to balance reconstruction accuracy with latent-space orthogonality.

\section{Methods}
\subsection{Primary problem setup}\label{sec:problem setup}
\begin{figure}[htb!]
    \centering
    \includegraphics[width=0.45\textwidth]{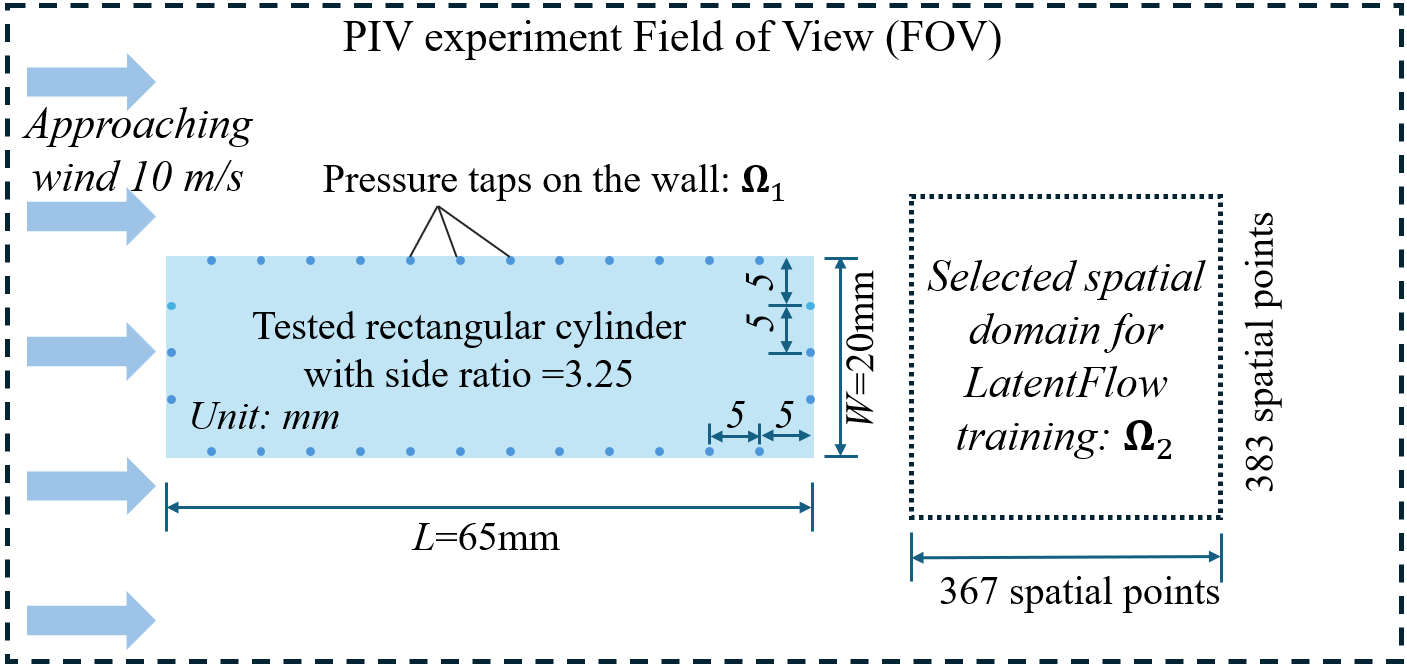}
    \vspace{-1ex}
    \caption{Spatial domains definition for the wind tunnel experiment. $\mathbf{\Omega}_1$ denotes sparse pressure taps on the wall, and $\mathbf{\Omega}_2$ denotes a selected wake domain.}
    \label{fig:piv}
\end{figure}

We denote the experimentally collected low-frequency (15~Hz) wall-pressure measurements as $\mathbf{P}_l$, which are measured at $n=30$ sparse locations in $\mathbf{\Omega}_1$. Simultaneously, synchronized low-frequency flow fields are collected as $\mathbf{U}_l \in \mathbb{R}^{N \times H \times W}$, where $N=2$ represents the number of flow variables (i.e., horizontal velocity $u$ and vertical velocity $v$), and $H=383$, $W=367$ denote the spatial points in the height and width directions within the domain $\mathbf{\Omega}_2$. High-frequency (512~Hz) wall-pressure measurements at the same locations in $\mathbf{\Omega}_1$ are denoted as $\mathbf{P}_h$.  
Figure~\ref{fig:piv} illustrates the cylinder dimensions and the domain definitions. Specifically, $\mathbf{\Omega}_1$ denotes the 30 sparse pressure taps on the wall, while $\mathbf{\Omega}_2$ consists of $383 \times 367$ spatial points, with 383 points distributed vertically and 367 points horizontally. 

Our objective is to develop a cross-modal framework, \emph{LatentFlow}, to reconstruct the high-frequency flow field $\mathbf{U}_h \in \mathbb{R}^{N \times H \times W}$ over $\mathbf{\Omega}_2$ from the high-frequency wall-pressure measurements $\mathbf{P}_h$. Mathematically, this corresponds to learning a unified mapping operator $\boldsymbol{\mathcal{M}}$:

\begin{equation}
    \boldsymbol{\mathcal{M}}: [\mathbf{P}_h ; \mathbf{U}_l, \mathbf{P}_l] \mapsto \mathbf{U}_h.
\end{equation}

\subsection{LatentFlow}\label{sec:latentflow}

\begin{figure*}[htb!]
    \centering
    \includegraphics[width=0.8\textwidth]{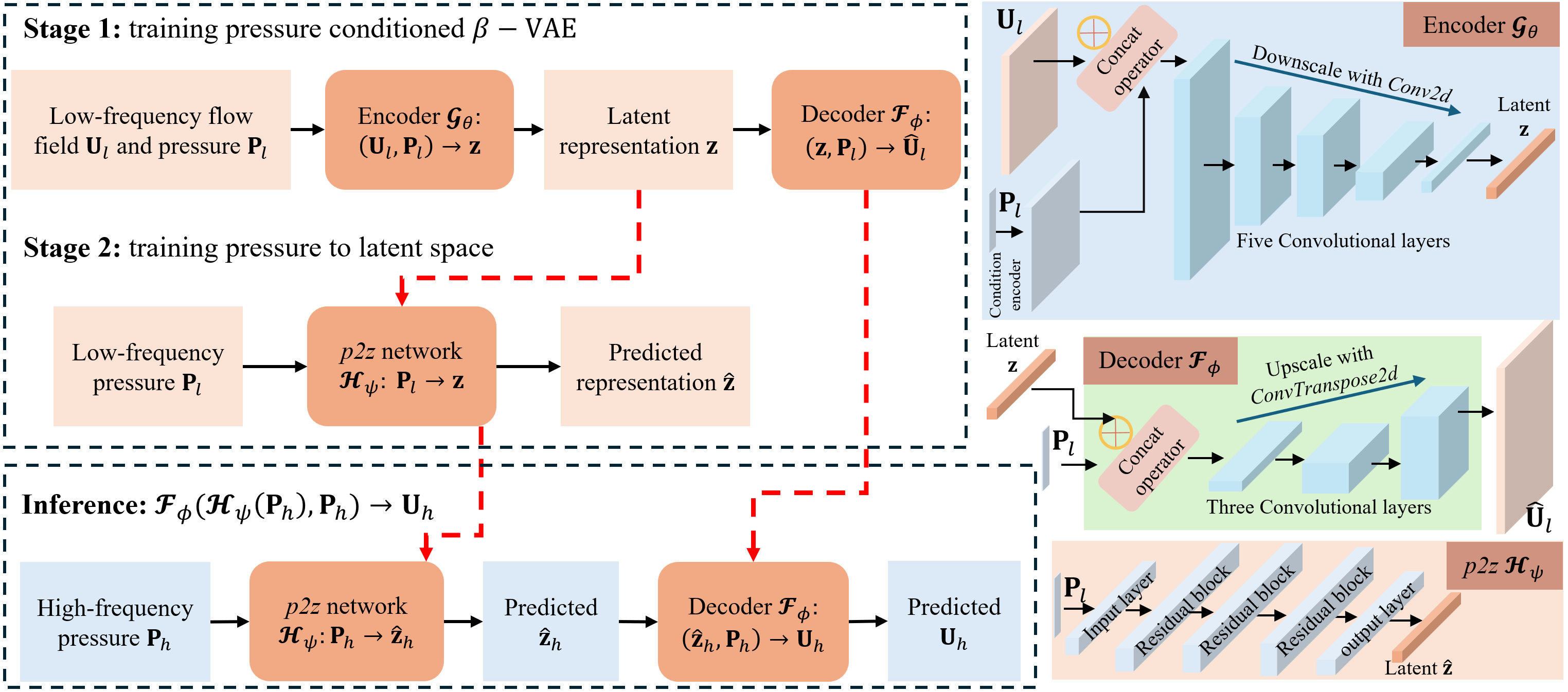}
    \vspace{-1ex}
    \caption{\textit{LatentFlow} details. The left dashed-line zone is the two-stage training and inference; the right colored-background zone is the component design. Stage 1: $p$C-$\beta$-VAE training via latent representation $\mathbf{z}$ based on low-frequency $\mathbf{U}_l$ and $\mathbf{P}_l$. Stage 2: learn $\boldsymbol{\mathcal{H}}_\psi$ from low-frequency pressure $\mathbf{P}_l$ to latent representation $\mathbf{z}$. In the inference stage,\textit{ LatentFlow} uses high-frequency pressure $\mathbf{P}_h$ for high-frequency flow field $\mathbf{U}_h$ reconstruction.
    }
    \label{fig: dataflow}
\end{figure*}

The pipeline and training strategy of the LatentFlow framework are presented in Figure~\ref{fig: dataflow}. In the first stage, we train a pressure-conditioned $\beta$-variational autoencoder ($p$C-$\beta$-VAE) to encode the low-frequency flow field $\mathbf{U}_l$ and the low-frequency wall pressure $\mathbf{P}_l$ into a latent representation $\mathbf{z}$, formulated as:

\begin{equation}
    \mathbf{z} = \boldsymbol{\mathcal{G}}_\theta (\mathbf{U}_l, \mathbf{P}_l),\quad \mathbf{z} \in \mathbb{R}^d,
    \label{eq:gtheta}
\end{equation}
where $\mathbf{z}$ is the $d$-dimensional latent variable capturing the coupled dynamics between $\mathbf{U}_l$ and $\mathbf{P}_l$.  
The decoder $\boldsymbol{\mathcal{F}}_\phi$ reconstructs the flow field $\widehat{\mathbf{U}}_l$ from the latent representation $\mathbf{z}$ and the simultaneous conditioning low-frequency pressure $\mathbf{P}_l$ as:

\begin{equation}
    \widehat{\mathbf{U}}_l = \boldsymbol{\mathcal{F}}_\phi(\mathbf{z}, \mathbf{P}_l).
    \label{eq:phi}
\end{equation}
In the $p$C-$\beta$-VAE model training, we set the total loss $\mathcal{L}_{\mathrm{VAE}}$ as follows:
\begin{align}
\mathcal{L}_{\mathrm{VAE}} 
&= \mathbb{E}_{\boldsymbol{\mathcal{G}}_\theta}
\left[ \lVert \mathbf{U}_l - \widehat{\mathbf{U}}_l \rVert^2_2 \right] \notag \\
&\quad + \beta \cdot D_{\mathrm{KL}}\left[ 
\boldsymbol{\mathcal{G}}_\theta \,\|\, p(\mathbf{z}) \right],
\label{eq:loss_cvae}
\end{align}
where the first term of $\mathcal{L}_{\mathrm{VAE}}$ is used to minimize the discrepancy between the reconstructed flow field $\widehat{\mathbf{U}}_l$ and the input ground truth flow field $\mathbf{U}_l$; the second term $D_{\text{KL}}[\cdot]$ is the Kullback–Leibler divergence \citep{kim2021comparing, asperti2020balancing, kullback1951information} that aims to adjust the distribution difference between the prior distribution $p(\mathbf{z})$ and the learned posterior distribution from $\boldsymbol{\mathcal{G}}_\theta$ in Eq.~\eqref{eq:gtheta}. For the prior distribution of $p(\mathbf{z})$, it follows the normal distribution as follows:
\begin{equation}
p(\mathbf{z}) = \mathbf{\mathcal{N}}(0, I).
\label{eq:pz}
\end{equation}
In Eq.~\eqref{eq:loss_cvae}, $\beta$ is a dynamic coefficient to adjust the discrepancy between two distributions as follows:
\begin{equation}
    \beta = \beta_{end}\frac{epo}{epochs},
    \label{eq:beta}
\end{equation}
where $\beta_{end}$ is the final coefficient for the variational $\beta$, $epo$ is the current training epoch and $epochs$ is the total training epochs. The training of $p$C-$\beta$-VAE model is aimed to get the hyperparameter $\theta$ in the encoder $\boldsymbol{\mathcal{G}}_\theta$ of Eq.~\eqref{eq:gtheta} and hyperparameter $\phi$ in the decoder $\boldsymbol{\mathcal{F}}_\phi$ of Eq.~\eqref{eq:phi} by minimizing the loss in Eq.~\eqref{eq:loss_cvae} as:

\begin{equation}
\theta, \phi = \arg \min_{\theta, \phi} \ \mathcal{L}_{\mathrm{VAE}}.
\end{equation}

In the second stage, we train a $p2z$ mapping from low-frequency wall pressure measurements 
$(\mathbf{P}_l)$ to the latent representation $\mathbf{z}$ learned in Stage~1 via encoder $\boldsymbol{\mathcal{G}}_\theta$ in Eq.~\eqref{eq:gtheta}. We define the $p2z$ network as $\boldsymbol{\mathcal{H}}_\psi$ mapping pressure to the latent representation parameterized by $\psi$:
\begin{equation}
    \hat{\mathbf{z}} = \boldsymbol{\mathcal{H}}_\psi(\mathbf{P}_l), 
    \quad \hat{\mathbf{z}} \in \mathbb{R}^d,
\end{equation}
where $\hat{\mathbf{z}}$ is the predicted latent representation distribution, which has the same dimension as $\mathbf{z}$ in Eq.~\eqref{eq:gtheta}.  
The $p2z$ network is trained by minimizing the mean squared error between the predicted latent vector $\hat{\mathbf{z}}$ and the ground truth latent vector $\mathbf{z}$ obtained from the Stage~1 encoder $\boldsymbol{\mathcal{G}}_\theta$. Besides, the Stage~1 trained decoder $\boldsymbol{\mathcal{F}}_\phi$ is frozen and utilized in the $p2z$ model training to perform end-to-end flow reconstruction. The $p2z$ model training loss is set as follows:
\begin{equation}
    \mathcal{L}_{{p2z}} = \frac{1}{d} \lVert \mathbf{z} - \hat{\mathbf{z}} \rVert_2^2 + \alpha  (\lVert \mathbf{U}_l - \widehat{\mathbf{U}}_l \rVert^2_2),
    \label{eq:loss_p2z}
\end{equation}
where $d=128$ is the latent dimension, a hyperparameter set for training. $\alpha$ is a dynamic factor for estimating flow reconstruction as calculated as follows:
\begin{equation}
    \alpha = \alpha_{end} \frac{epo}{epos_{p2z}},
    \label{eq:alpha}
\end{equation}
where $\alpha_{end}$ is the final coefficient, $epo$ is the current training epoch, and $epos_{p2z}$ is the total training epochs of the $p2z$ model. The loss design in Eq.~\eqref{eq:loss_p2z} aims to guide the model not only in learning the mean characteristic of $\mathbf{z}$, but also in predicting the flow field variation to guide high-frequency flow field reconstruction. Once trained, $\boldsymbol{\mathcal{H}}_\psi$ can be used together with the Stage~1 decoder $\boldsymbol{\mathcal{F}}_\phi$ 
to reconstruct the high-frequency flow field from high-frequency wall pressure $\mathbf{P}_h$:
\begin{equation}
    {\mathbf{U}}_h = \boldsymbol{\mathcal{F}}_\phi(\boldsymbol{\mathcal{H}}_\psi(\mathbf{P}_h), \mathbf{P}_h).
\end{equation}

\section{Experiments}
\subsection{Dataset}
\begin{figure}[htb!]
    \centering
    \includegraphics[width=0.42\textwidth]{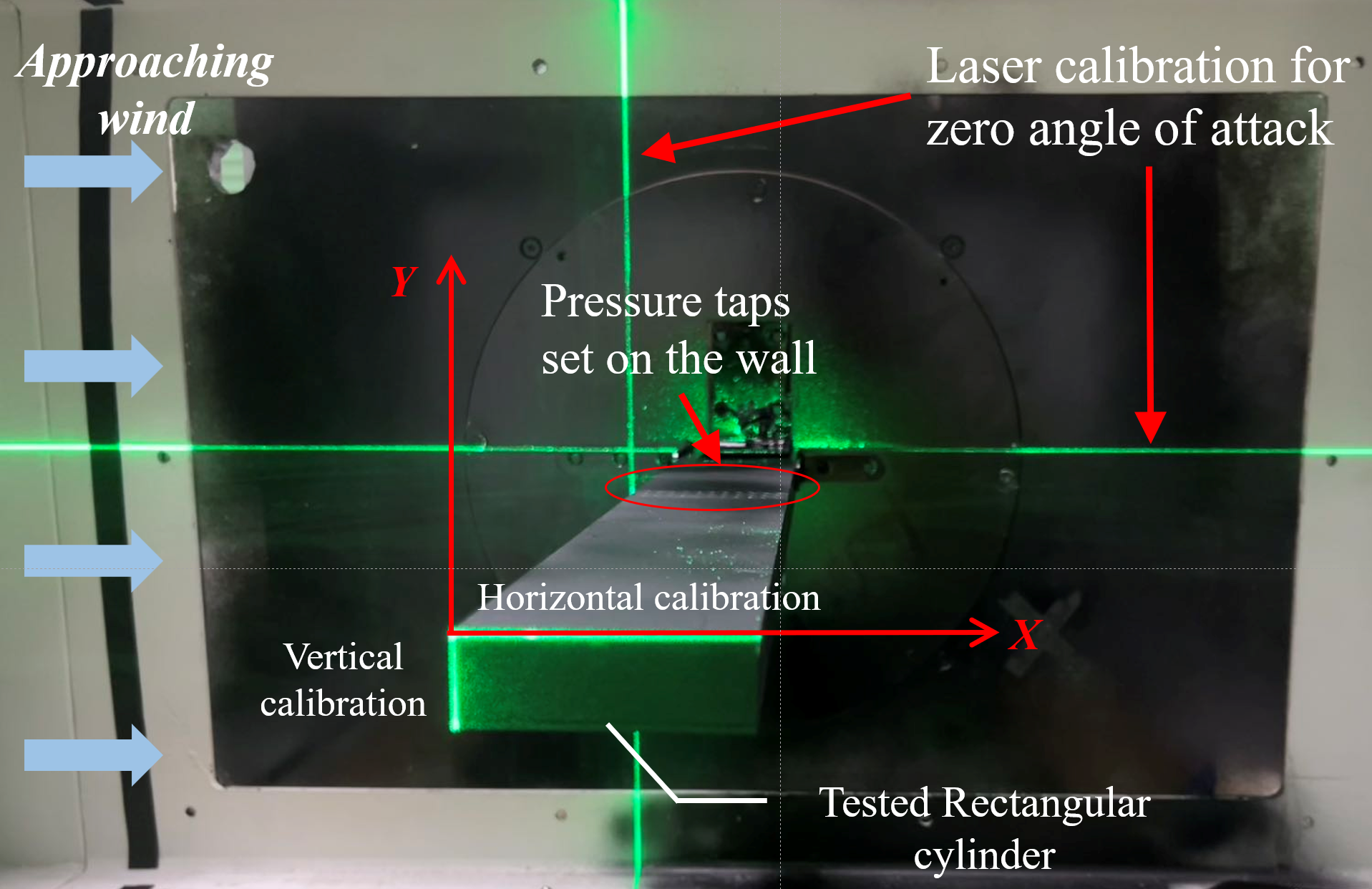}
    \vspace{-1ex}
    \caption{Front view of the Wind tunnel test: rectangular cylinder is set on the rotating end-plate, wall pressure measurements are set on the surface, green lines denote the zero angle of attack calibration via a 12-line laser instrument.}
    \label{fig:experiment}
\end{figure}

The dataset was obtained from wind tunnel experiments on a rectangular cylinder with a side ratio of 3.25 under zero angle of attack in the \textit{Artificial Intelligence for Wind Engineering} laboratory at Harbin Institute of Technology, Shenzhen. Figure~\ref{fig:experiment} shows the experimental layout. Two types of data were collected: (1) low-frequency (15~Hz) wake flow fields $\mathbf{U}_l$ over domain $\mathbf{\Omega}_2$ and wall-pressure measurements $\mathbf{P}_l$ over domain $\mathbf{\Omega}_1$, totaling 1,200 snapshots; (2) high-frequency (512~Hz) wall-pressure measurements $\mathbf{P}_h$ over $\mathbf{\Omega}_1$, totaling 30,000 snapshots. 
Both types of data were acquired in the same wind tunnel experiment, differing only in sampling frequency. Experimental setup details are described in \citet{liu2025aerodynamic};  statistical verification and physical interpretation are provided in \citet{liu2024exploring}. More details about PIV processing are expressed in Appendix A. After collection, the flow fields in the wake domain $\mathbf{\Omega}_2$ were normalized to a standard normal distribution $\mathcal{N}_v(0, 1)$. The pressure measurements are converted into pressure coefficients $C_p$ defined as $C_p = p / (0.5 \rho u^2)$, where $\rho = 1.225~\mathrm{kg/m^3}$ is the air density and $u = 10~\mathrm{m/s}$ is the incoming flow velocity. 

For the synchronized low-frequency flow fields and wall-pressure measurements, 90\% of the temporal snapshots are used to train the $p$C-$\beta$-VAE model, while the remaining 10\% serve for model evaluation in terms of both instantaneous and time-averaged flow fields. During inference, only the high-frequency wall-pressure measurements $\mathbf{P}_h$ are used to reconstruct the high-frequency wake flow field. To evaluate high-frequency reconstructed flow field data, we synchronize the high-frequency instantaneous flow field with the low-frequency experimental instantaneous flow field (ground truth) by matching the corresponding lift coefficient $C_l$. More details on synchronizing the inferred high-frequency flow field to low-frequency experimental ground truth are presented in Appendix B. Besides, the predicted high-frequency fields are assessed through statistical metrics and physical analyses.

\subsection{Implementation Details}
\subsubsection{Model Design}
Figure~\ref{fig: dataflow} illustrates the \textit{LatentFlow} pipeline and architecture. To achieve robust high-frequency flow field reconstruction, we carefully design the number of layers and neurons for each component. The architecture includes: (1) the encoder and decoder of the $p$C-$\beta$-VAE model, responsible for mapping between flow fields and latent space as well as reconstruction; and (2) the $p2z$ network, which projects sparse wall-pressure measurements to the latent space. In the encoder of $p$C-$\beta$-VAE model, we use 5 layers of convolutional layers to downscale the flow field to a latent representation. In the decoder part, three deconvolutional layers are used to upscale the latent representation. In the $p2z$ model, we set an input layer, three residual block layers, and an output layer. In the residual block, three linear projection layers are used, and $dropout$ as well as $batchnorm$ are applied to enhance the $p2z$ model robustness. We explored various network depths, latent dimensions, and residual block configurations (more details in Appendix C).

\subsubsection{Model Training}
Training is performed in two stages, with parameters listed in Table~\ref{tab:training parameters}. In Stage 1, the $p$C-$\beta$-VAE is trained to obtain the encoder $\boldsymbol{\mathcal{G}}_\theta$ and decoder $\boldsymbol{\mathcal{F}}_\phi$. Stage 2 uses a special training strategy to train the $p2z$ network $\boldsymbol{\mathcal{H}}_\psi$. Specifically, instantaneous sparse low-frequency wall-pressure measurements $\mathbf{P}_l$ are input to $p2z$ to predict latent codes $\hat{\mathbf{z}}$ by minimizing discrepancy between $\boldsymbol{\mathcal{G}}_\theta$ outputs and $\hat{\mathbf{z}}$. During this stage, the decoder $\boldsymbol{\mathcal{F}}_\phi$ is frozen and used to compute the end-to-end flow field reconstruction loss as defined in Eq.~\eqref{eq:loss_p2z}. The \textit{LatentFlow} model is trained with an \textit{A40 Navidia GPU} at Hong Kong University of Science and Technology, Department of Civil and Environmental Engineering, with four CPU cores used for data transition. 

\begin{table}[htb!]
    \centering
    \vspace{-1ex}
    \caption{Training parameters for \textit{LatentFlow}.}
    \vspace{-1ex}
    \label{tab:training parameters}
    \begin{tabular}{cc}
    \hline
    Hyperparameter & Settings \\
    \hline
       Learning rate ($p$C-$\beta$-$\mathrm{VAE}$)  & $10^{-4}$  \\
       Learning rate ($p2z$)   & $10^{-3}$ \\
       Batch size ($p$C-$\beta$-$\mathrm{VAE}$) & 4\\
       Batch size ($p2z$) & 2\\
       Latent Dimension ($\mathbf{z}$) in Eq.~\eqref{eq:gtheta},~\eqref{eq:phi},~\eqref{eq:loss_cvae},~\&\eqref{eq:pz} & 128 \\
       $\beta_{end}$ in Eq.~\eqref{eq:beta} & 0.001 \\
       $epochs$ in Eq.\eqref{eq:beta} & 2000\\
       $epos_{p2z}$ in Eq.~\eqref{eq:alpha} & 1000 \\
       $\alpha_{end}$ & 0.01\\
       Dropout rate in $p2z$ network ($\boldsymbol{\mathcal{H}}_\psi$) & 0.2 \\
       
       \hline
    \end{tabular}
\end{table}

\subsection{Results and evaluations}

As described in the problem setup, the ultimate goal is to reconstruct high-frequency wake flow fields from high-frequency wall-pressure inputs. The \textit{LatentFlow} framework comprises two training stages ($p$C-$\beta$-VAE and $p2z$) followed by an inference stage. Accordingly, model performance is evaluated from two aspects: (1) $p$C-$\beta$-VAE flow reconstruction, and (2) high-frequency flow inference.

\subsubsection{$p$C-$\beta$-VAE evaluation}
\begin{figure}[htb!]
    \centering
    \vspace{-3ex}
    \includegraphics[width=0.47\textwidth]{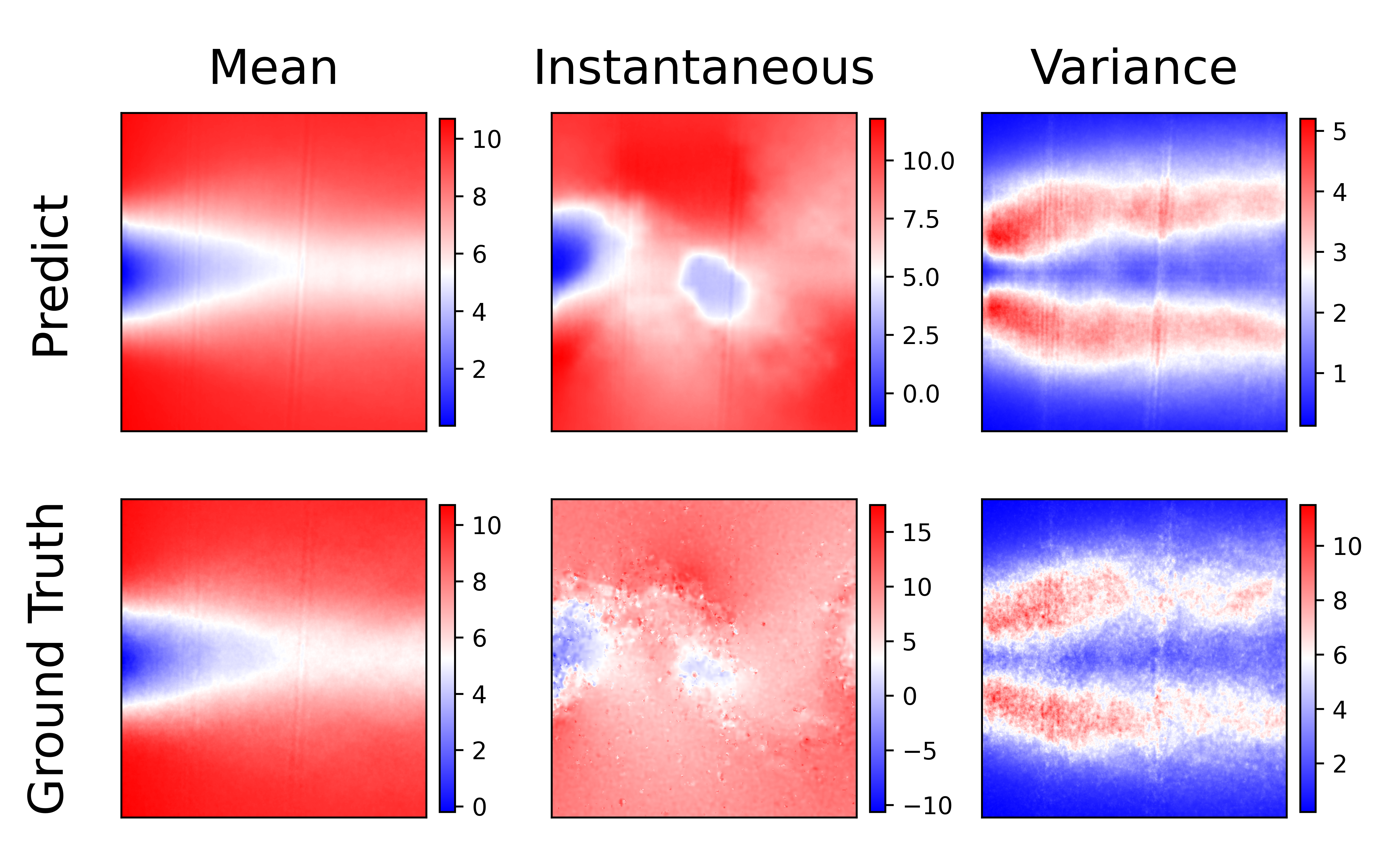}
    \vspace{-4ex}
    \caption{Time-averaged flow field $\bar{u}$, instantaneous flow field $u$, and flow field $u$ variance in temporal sequence comparison between $p$C-$\beta$-VAE reconstructed data and low-frequency experimental ground truth. }
    \label{fig:instantaneous_compare}
\end{figure}
For the $p$C-$\beta$-VAE stage evaluation, 10\% of the low-frequency dataset (120 snapshots), low-frequency wall-pressure measurements $\mathbf{P}_l$ are input to the trained $p$C-$\beta$-VAE model to generate reconstructed flow fields. The predicted wake flow fields are evaluated by comparing instantaneous velocity fields, time-averaged flow fields, and temporal variances against experimental measurements. As shown in Figure~\ref{fig:instantaneous_compare}, the predicted time-averaged velocity field $\bar{u}$ closely matches the experimental ground truth, indicating that the $p$C-$\beta$-VAE model effectively captures the mean flow characteristics. For instantaneous flow fields, experimental data exhibit small-scale structures and large gradients in the wake vortices. While the $p$C-$\beta$-VAE model successfully captures the wake vortices, it demonstrates limited capability in resolving finer vortex structures. These evaluations allow verification of the model's ability to capture both mean flow features and unsteady dynamics, which further lay a foundation for\textit{ LatentFlow} to infer high-frequency turbulent flow.

\subsubsection{\textit{LatentFlow} inference evaluation}
\begin{figure}[htb!]
    \centering
    \subfigure[]{\includegraphics[width=0.4\textwidth]{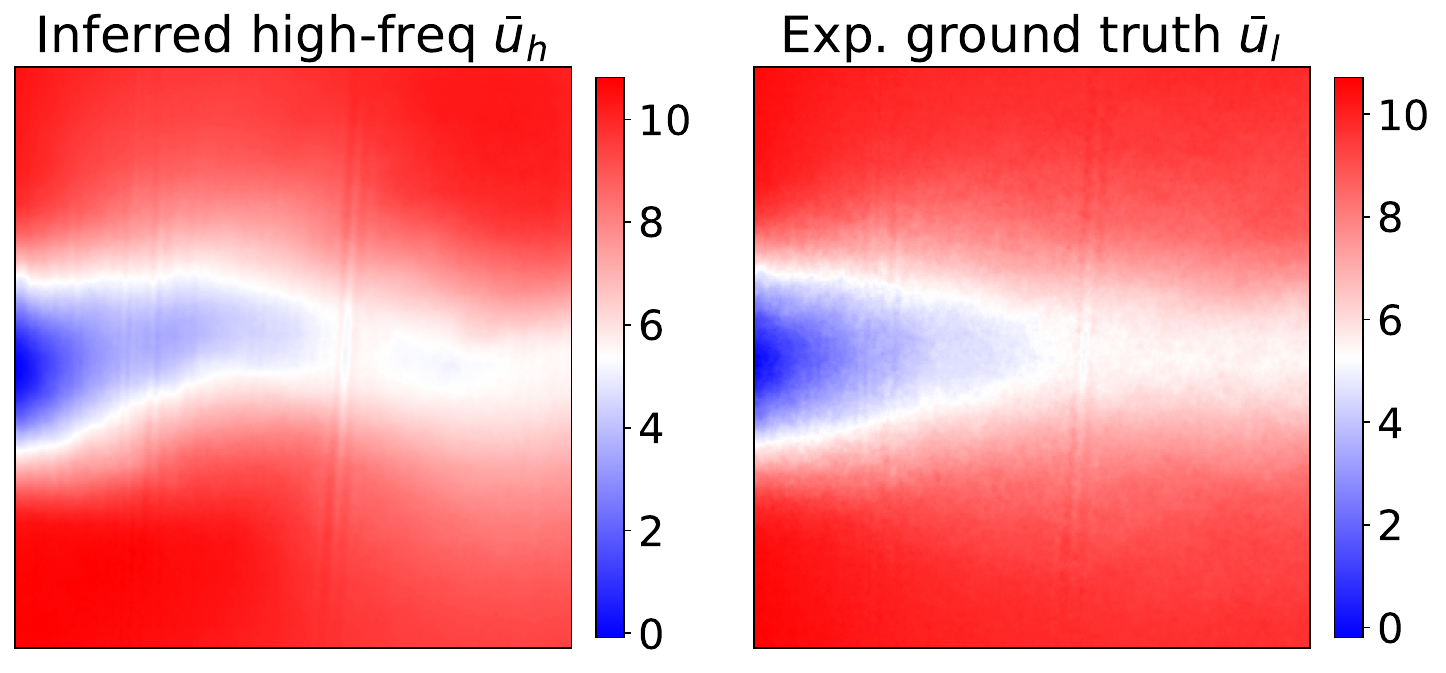}\label{fig:infer_mean_flow_compare}}
    \hfill
    \subfigure[]{\includegraphics[width=0.4\textwidth]{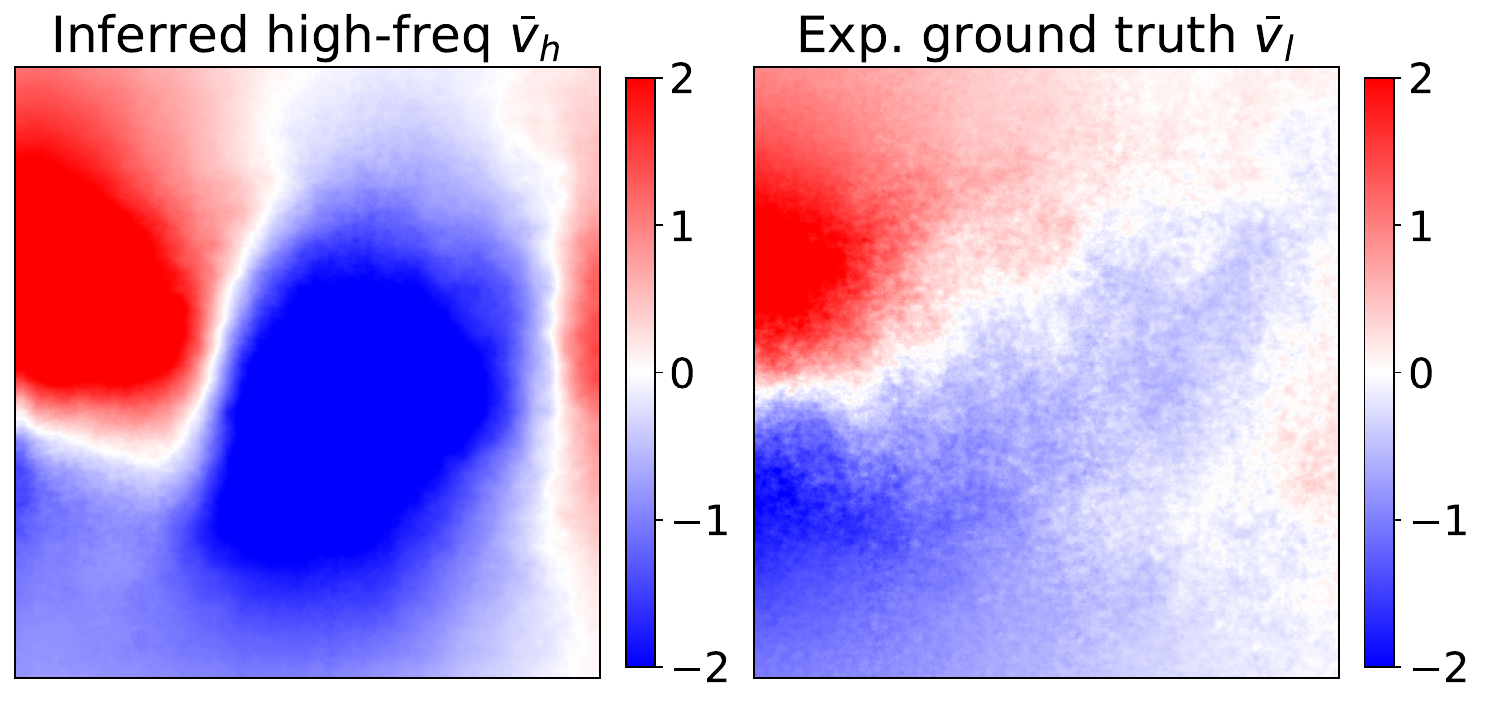}\label{fig:infer_mean_v}}
    \vspace{-2ex}
    \caption{Time-averaged flow field comparison: (a). \textit{LatentFlow} inferred horizontal velocity $\bar{u}_h$ and experimental ground truth $\bar{u}_l$. (b) \textit{LatentFlow} inferred vertical velocity $\bar{v}_h$ and experimental ground truth $\bar{v}_l$.}
    \label{fig:mean compare}
\end{figure}

During inference, our goal is to predict high-frequency wake flow fields $\mathbf{U}_h$ using the trained \textit{LatentFlow} model. Therefore, we use 1,024 snapshots of high-frequency wall pressure $\mathbf{P}_h$ as input, i.e., 2~s in the physical time domain, and the corresponding 1,024 snapshots of predicted high-frequency $\mathbf{U}_h$ are collected from the \textit{LatentFlow}. Similar to $p$C-$\beta$-VAE evaluation, we compare the time-averaged high-frequency flow field $\overline{u}_h$ and $\overline{v}_h$ with the experimental time-averaged low-frequency flow field $\overline{u}_l$ and $\overline{v}_l$ shown in Figures~\ref{fig:infer_mean_flow_compare} and~\ref{fig:infer_mean_v}. We found that the inferred time-averaged high-frequency flow field $\overline{u}_h$ presents minor differences in the middle part of the wake compared to the experimental low-frequency flow field $\overline{u}_l$. Although for the inferred $\bar{v}_h$, the negative velocity zone of the wake is more concentrated, the experimental $\bar{v}_l$ is smoother in the wake shown in Figure~\ref{fig:infer_mean_v}. 

Besides, we evaluate \textit{LatentFlow} by matching the inferred instantaneous flow field with the low-frequency experimental flow field through the time-history lift coefficient (details in Appendix B). Figures~\ref{fig:inferred_u_instant} and~\ref{fig:instant_v} compare the instantaneous velocity field between the inferred high-frequency prediction and the experimental low-frequency ground truth at $C_l = C_{l,\max}$. The results show that \textit{LatentFlow} successfully captures the overall flow pattern, including the von K\'arm\'an vortex street \citep{davis1982numerical, heil2017topological} shown in Figure~\ref{fig:inferred_u_instant}. However, fine-scale vortical structures and regions with strong gradients are not fully resolved, which is consistent with the observations from the $p$C-$\beta$-VAE evaluation (Figure \ref{fig:instantaneous_compare}). This limitation can be attributed to both experimental conditions and the intrinsic characteristics of turbulent flows. As the incoming flow interacts with the bluff body, the turbulent wake exhibits periodic vortex shedding with numerous small-scale vortices responsible for energy dissipation, thereby forming the strong gradient zones observed in the experimental ground truth.
\begin{figure}[htb!]
    \centering
    \subfigure[]{\includegraphics[width=0.42\textwidth]{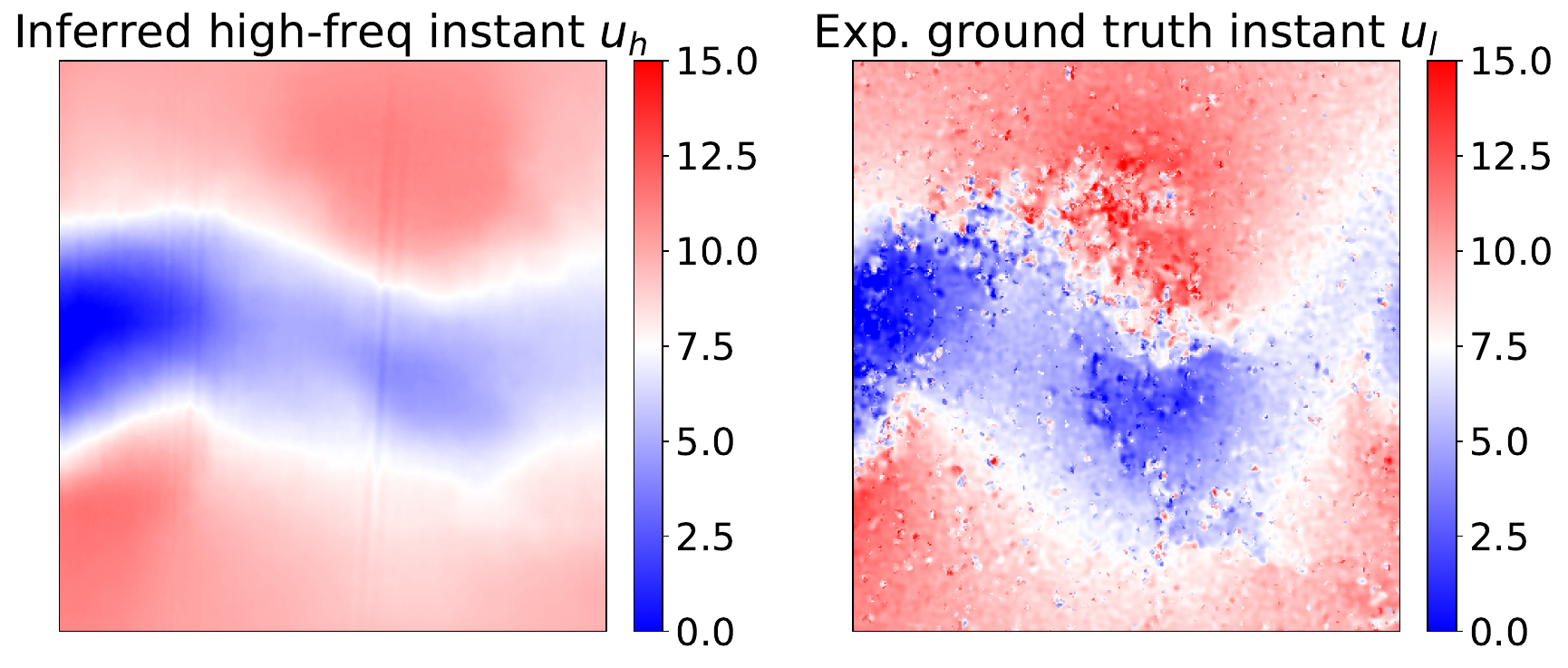}\label{fig:inferred_u_instant}}
    \subfigure[]{\includegraphics[width=0.42\textwidth]{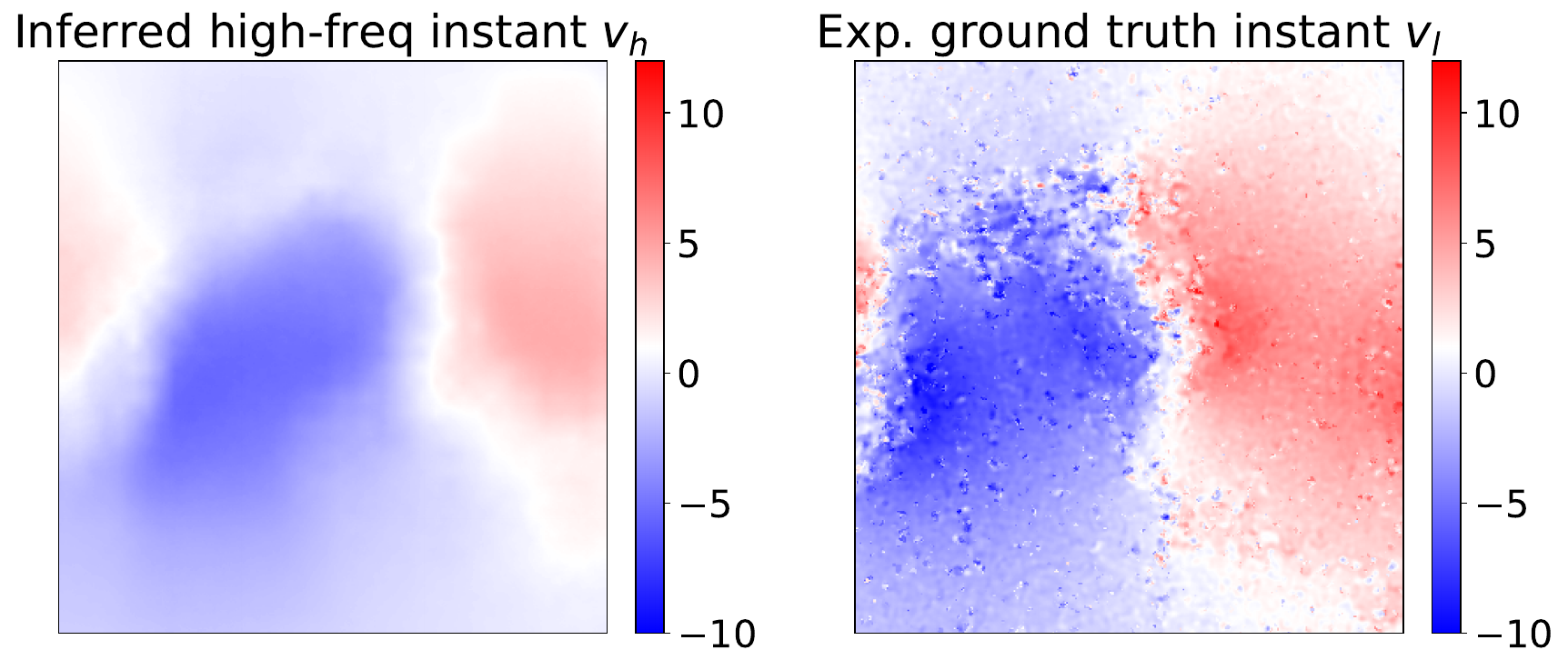}\label{fig:instant_v}}
    \vspace{-3ex}
    \caption{Instantaneous flow field comparison at $C_l = C_{l,\max}$: (a) \textit{LatentFlow} inferred high-frequency instantaneous horizontal velocity $u_h$ and experimental ground truth $u_l$, (b) \textit{LatentFlow} inferred high-frequency instantaneous vertical velocity $v_h$ and experimental ground truth $v_l$. }
\end{figure}

\begin{figure}[htb!]
    \centering
    \includegraphics[width=0.43\textwidth]{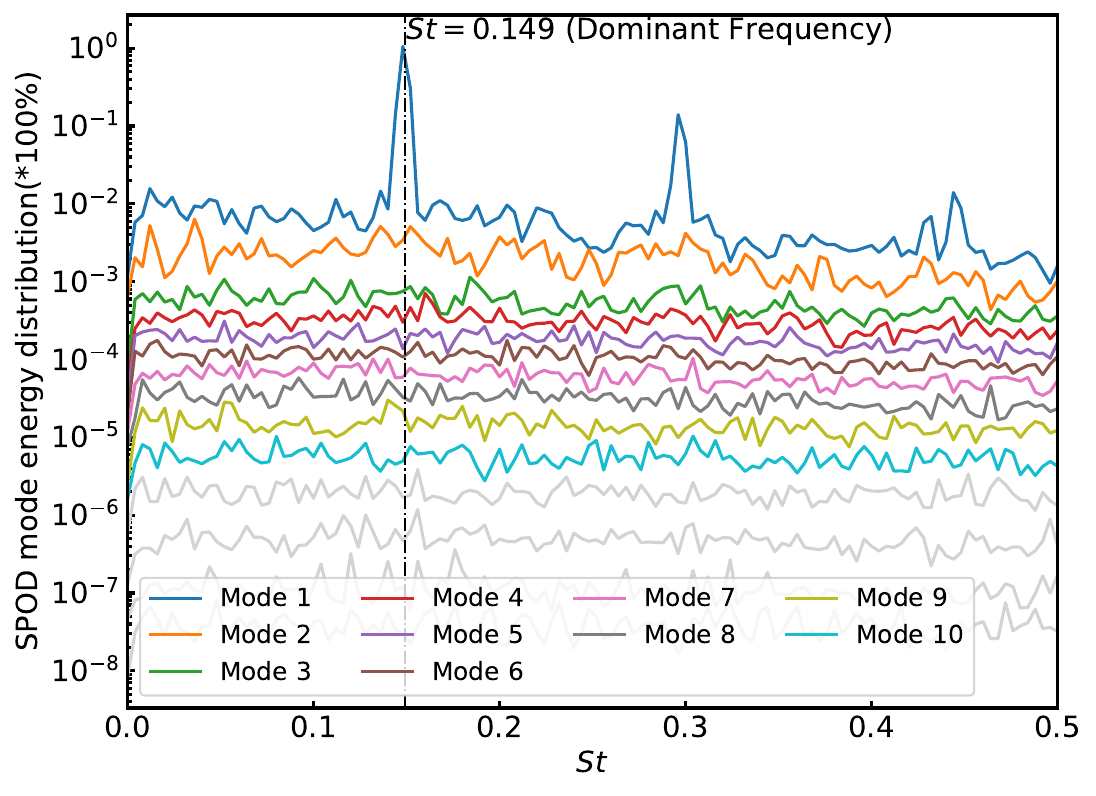}
    \vspace{-2ex}
    \caption{ SPOD energy distribution of inferred high-frequency flow field $u_h$. The $x-$ axis is the Strohaul number ($St=fD/U$), calcualted from frequency $f$, rectangular cylinder width $D=0.02~m$, and approaching wind velocity $U=10~m/s$.}
    \label{fig:energy_dsitribution}
\end{figure}

Furthermore, we analyzed the high-frequency flow field physical interpretation based on dominant modes and frequency characteristics using the spectral proper orthogonal decomposition (SPOD) technique \citep{sieber2016spectral, towne2018spectral}, which is widely adopted for dynamical system characterization. The theoretical and implementation details of SPOD are expressed in Appendix D. As shown in Figure~\ref{fig:energy_dsitribution}, SPOD captures the dominant frequency (74.5~Hz), twice the dominant frequency (149~Hz), and three times the dominant frequency ($\approx$ 225~Hz) from the inferred high-frequency results as presented with three peaks. These frequency characteristics are in the inferred high-frequency flow field and correspond to the existing work \citep{liu2024exploring}. However, these high-frequency characteristics are not captured by low-frequency (15~Hz) experimental data because of low-frequency sampling. In other words, from the above validation, the \textit{LatentFlow} can predict the high-frequency flow field physical information based on low-frequency data training, which eventually makes cross-frequency inference.

\section{Conclusion and outlook}
In this work, we proposed \textit{LatentFlow} to infer experimental high-frequency flow field based on high-frequency sparse pressure measurements via latent mapping. The \textit{LatentFlow} framework is first trained via a $p$C-$\beta$-VAE network by encoding synchronized low-frequency flow field and sparse wall pressure to a latent representation. The decoder is trained on the encoded latent representation conditioned on the wall pressure measurements. Furthermore, we train the $p2z$ network to project the low-frequency wall pressure to the latent representation, which is the output of $p$C-$\beta$-VAE. In the inference stage, the high-frequency sparse wall pressure is set as input to the \textit{LatentFlow} framework, and the model predicts high-frequency flow field. Statistical investigation and physical interpretation are performed to evaluate the \textit{LatentFlow} performance, showing that \textit{LatentFlow} is capable of predicting high-frequency flow field and mining the physics from the limited low-frequency data. 

There are some future directions to improve this work. As \textit{ LatentFlow }framework depends on latent representation to reconstruct turbulent wake flow, indicating that the trained latent representation is essential in the model. The current latent framework is trained on the basis of limited experimental data. In the future, data fusion techniques can be considered by integrating both experimental data and numerical data to generalize the model. Besides, the trained \textit{LatentFlow} model is purely dependent on instantaneous information, while some time-series physical boundaries can be potentially embedded into the framework to enhance the model robustness.

\section{Acknowledgment}
J.L. and K.T., acknowledge support from the Hong Kong Research Grants Council (RGC) General Research Fund (Project No. 16211821). G.H. acknowledges support from the National Natural Science Foundation of China (Project No. 52441803) and the Aero Science Foundation of China‌‌ (2024M034077001). 

\newpage
\appendix
\section{Appendix A: PIV data processing}

Following the wind tunnel experiments, the low-frequency velocity fields $\mathbf{U}_l$ and sparse wall-pressure measurements $\mathbf{P}_l$ were obtained. The velocity fields around the rectangular cylinder were recorded by a high-speed camera and processed using the \textit{Davis} software package.

\subsection*{Pre-processing}
Prior to the particle image velocimetry (PIV) computation, several pre-processing steps were applied to enhance data quality:
\begin{itemize}
    \item \textbf{Background removal:} The \textit{Time Series Group – Subtract Time Filter} was applied to suppress background noise in the recorded images.
    \item \textbf{Noise reduction:} A \textit{Gaussian average} filter was applied to each set of three consecutive frames to further reduce high-frequency noise.
    \item \textbf{Masking:} A \textit{Mask Group – Geometric Mask} was used to exclude the solid body region (rectangular cylinder) from the field of view (FOV).
\end{itemize}

\subsection*{PIV processing parameters}
The PIV analysis was carried out in standard PIV mode with perspective correction enabled. The following parameters were used:
\begin{itemize}
    \item Maximum expected particle displacement: 12 pixels.
    \item Initial interrogation window: $48 \times 48$ pixels, rectangular weighting, 50\% overlap.
    \item Final interrogation window: $12 \times 12$ pixels, circular weighting, 75\% overlap.
    \item Final spatial resolution: $0.86\,\mathrm{mm} \times 0.86\,\mathrm{mm}$ in the physical domain.
\end{itemize}

Based on the above parameters, the PIV algorithm provided instantaneous velocity fields $\{u_i(X,Y), v_i(X,Y)\}$ for the $i$-th snapshot. 

\subsection*{Post-processing}
For this case, 1,200 instantaneous snapshots were obtained. The total processing time was approximately 4 hours on a workstation equipped with an 11th Gen Intel(R) Core(TM) i7-11700 CPU @ 2.50 GHz. Time-averaged velocity fields were calculated using the \textit{Statistical Group – Vector Statistics} function in \textit{Davis}:
\begin{equation}
    \bar{u}(X,Y) = \frac{1}{t} \sum_{j=1}^{t} u_j(X,Y),
\end{equation}
where $t=1200$ in this study. Similar expressions were used for $\bar{v}(X,Y)$.

\section{Appendix B: Instantaneous wake flow match with lift coefficient}
\begin{figure}[htb!]
    \centering
    \includegraphics[width=0.47\textwidth]{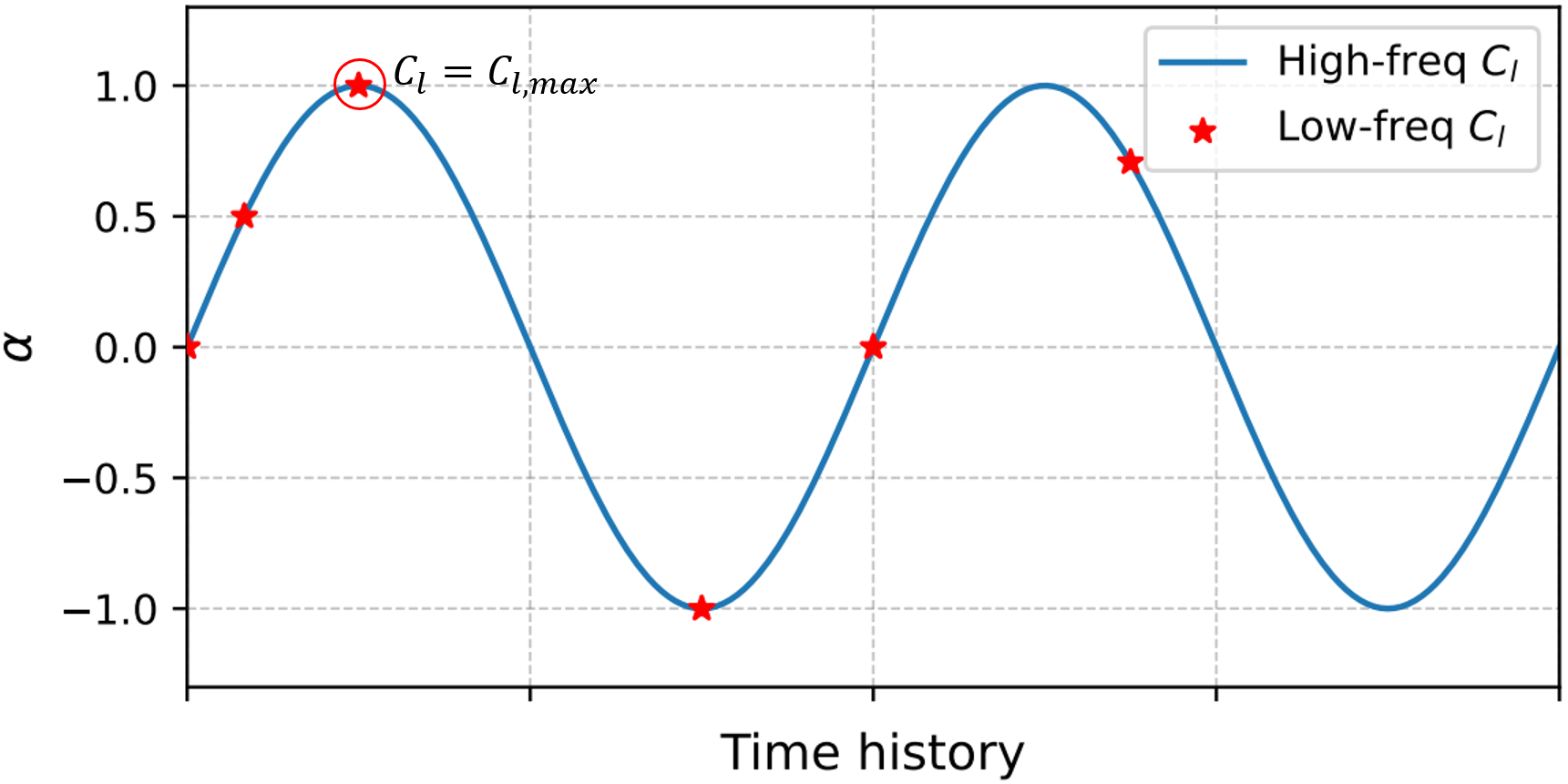}
    \caption{Lift coefficient time history for instantaneous information match: low-frequency instantaneous information is a sub-sampling of high-frequency information.}
    \label{fig:cl}
\end{figure}

We collected two types of wall pressure measurements as low-frequency (15~Hz) $\mathbf{P}_l$ and high-frequency (512~Hz)  $\mathbf{P}_h$. The lift coefficient can be calculated as $C_{l,\mathrm{low}}$ and $C_{l,\mathrm{high}}$ from the low-frequency and high-frequency pressure measurements, respectively. The high-frequency lift coefficient is a periodic and continuous $sin$ function of time history. The low-frequency data is the sub-sampling of the dynamic system, treated as scatter points on the $sin$ line. Two datasets share the same lift coefficient history statistical information, as two types of data are sampled from an identical wind tunnel test, apart from the sampling frequency. Figure~\ref{fig:cl} shows that low-frequency information is a sub-sampling of the high-frequency lift coefficient history. Based on this lift coefficient sharing, we match the high-frequency lift coefficient snapshot to the low-frequency lift coefficient snapshot at $C_{l,\mathrm{high}}=C_{{l},\mathrm{max}}=C_{l,\mathrm{low}}$. At this instantaneous snapshot, the experimental ground truth flow field is available, which can be used for high-frequency flow field comparison. 
\section{Appendix C: Model parameters and layer design}
\textit{LatentFlow} comprises two main components: the $p$C-$\beta$-VAE and the $p2z$ model. For the $p$C-$\beta$-VAE, we varied the number of downscale/upscale layers and the latent representation dimensions, which directly influence reconstruction accuracy. For the $p2z$ model, we compared three architectures: (i) standard MLP, (ii) MLP with residual connections, and (iii) MLP with residual connections plus dropout. We also evaluated different numbers of residual blocks in the $p2z$ model. Due to the page limit in the main text, only the selected satisfactory configurations are reported here.
\section{Appendix D: Spectral proper orthogonal decomposition (SPOD)}

SPOD is used to estimate the inferred high-frequency flow field. SPOD is based on proper orthogonal decomposition (POD) theory. 

\subsubsection{POD Theory}

Consider a high-frequency time series of flow-field snapshots 
$\boldsymbol{Q}\in \mathbb{R}^{t\times H\times W}$ over $t$ instants on a spatial grid of size $H\times W$. 
Let $\boldsymbol{q}_j \in \mathbb{R}^{H\times W}$ denote the $j$-th snapshot and define
\begin{equation}
    \boldsymbol{Q} = [\boldsymbol{q}_1,\,\boldsymbol{q}_2,\,\ldots,\,\boldsymbol{q}_t], 
\end{equation} where $t=1024$ is the total time instants for this work. 
The time-averaged field is
\begin{equation}
    \bar{\boldsymbol{q}} = \frac{1}{t}\sum_{j=1}^{t} \boldsymbol{q}_j \in \mathbb{R}^{H\times W}.
\end{equation}
Define the fluctuation snapshots $\tilde{\boldsymbol{q}}_j = \boldsymbol{q}_j - \bar{\boldsymbol{q}}$ and vectorize each snapshot as 
$\tilde{\boldsymbol{x}}_j = \mathrm{vec}(\tilde{\boldsymbol{q}}_j)\in\mathbb{R}^{M}$ with $M=H*W$. 
Stacking them in time yields the data matrix
\begin{equation}
    \hat{\boldsymbol{Q}} = 
    \begin{bmatrix}
        \tilde{\boldsymbol{x}}_1^\top\\
        \tilde{\boldsymbol{x}}_2^\top\\
        \vdots\\
        \tilde{\boldsymbol{x}}_t^\top
    \end{bmatrix}
    \in \mathbb{R}^{t\times M}.
\end{equation}
The spatial covariance matrix is then calculated as follows:
\begin{equation}
    \boldsymbol{C}=\frac{1}{t-1}\,\hat{\boldsymbol{Q}}^\top \hat{\boldsymbol{Q}}\in\mathbb{R}^{M\times M}. 
    \label{eq:pod_cov}
\end{equation}
Solving the eigenvalue problem for the above covariance matrix $\boldsymbol{C}$
\begin{equation}
    \boldsymbol{C}\,\boldsymbol{\Phi}_i = \lambda_i\,\boldsymbol{\Phi}_i,
\end{equation}
yields POD spatial modes $\boldsymbol{\Phi}_i\in\mathbb{R}^{M}$ and eigenvalues $\lambda_i\ge 0$, with orthonormality:
\begin{equation}
    \langle \boldsymbol{\Phi}_i,\boldsymbol{\Phi}_j\rangle = \delta_{ij}.
\end{equation}
The temporal coefficients for each snapshot are
\begin{equation}
    a_{i j} = \boldsymbol{\Phi}_i^\top \tilde{\boldsymbol{x}}_j, 
    \qquad \text{equivalently } \ \boldsymbol{A}=\hat{\boldsymbol{Q}}\,\boldsymbol{\Phi}\ \in \mathbb{R}^{t\times r},
\end{equation}
where $\boldsymbol{\Phi}=[\boldsymbol{\Phi}_1,\ldots,\boldsymbol{\Phi}_r]$ and $r=\mathrm{rank}(\hat{\boldsymbol{Q}})$.
A $k$-mode POD approximation of the fluctuation field at time $j$ is
\begin{equation}
    \tilde{\boldsymbol{x}}_j \approx \sum_{i=1}^{k} a_{i j}\,\boldsymbol{\Phi}_i.
\end{equation}
Reshaping back to the grid gives $\tilde{\boldsymbol{q}}_j\in\mathbb{R}^{H\times W}$, and the full-field reconstruction is
\begin{equation}
    \boldsymbol{q}_j \approx \bar{\boldsymbol{q}} + \sum_{i=1}^{k} a_{i j}\,\mathrm{unvec}(\boldsymbol{\Phi}_i).
\end{equation}
The cumulative captured energy is 
\begin{equation}
    \mathcal{E}_k=\frac{\sum_{i=1}^{k}\lambda_i}{\sum_{i=1}^{r}\lambda_i}.
\end{equation}

\subsubsection{SPOD Theory}

Unlike the space-only POD, spectral proper orthogonal decomposition (SPOD) identifies modes that are coherent in both space and time \citep{towne2018spectral}.  Let $\tau = k\,\Delta t$ be the temporal separation between snapshots.  
The space–time correlation (cross-covariance) matrix is defined as
\begin{equation}
    \boldsymbol{C}(\tau) = \frac{1}{t-k} \sum_{j=1}^{t-k} 
    \tilde{\boldsymbol{x}}_j \, \tilde{\boldsymbol{x}}_{j+k}^\top
    \ \in\mathbb{R}^{M\times M}, 
\end{equation}
where $\tilde{\boldsymbol{x}}_j = \mathrm{vec}(\tilde{\boldsymbol{q}}_j)$ is the vectorized fluctuation field at time $j$, and $M = H*W$.

The spectral density matrix is obtained by the Fourier transform of the correlation matrix:
\begin{equation}
    \boldsymbol{S}(f) = \sum_{k=-\infty}^{\infty} 
    \boldsymbol{C}(k\Delta t) \, e^{- i 2\pi f\, k \Delta t}.
    \label{eq:spod_S}
\end{equation}
In practice, $\boldsymbol{S}(f)$ is estimated using Welch's method: the time series is divided into $n_b=14$ blocks of length $n_{\mathrm{Dft}}=256$ with $overlap=200$, each block is Fourier transformed in time, and the resulting Fourier coefficients are used to form the cross-spectral density (CSD) matrix at each frequency.

At each frequency $f$, the SPOD modes are obtained from the Hermitian eigenvalue problem
\begin{equation}
    \boldsymbol{S}(f) \, \boldsymbol{\Psi}_i(f)
    = \lambda_i(f) \, \boldsymbol{\Psi}_i(f),
\end{equation}
where $\boldsymbol{\Psi}_i(f) \in \mathbb{R}^{M}$ is the $i$-th SPOD mode at frequency $f$, normalized such that
\begin{equation}
    \langle \boldsymbol{\Psi}_i(f), \boldsymbol{\Psi}_j(f) \rangle
    = \delta_{ij},
\end{equation}
and $\lambda_i(f) \ge 0$ is the corresponding modal spectral energy.

The correlation function can be reconstructed from the spectral decomposition as
\begin{equation}
    \boldsymbol{C}(\tau) =
    \int_{-\infty}^{\infty} \sum_{i=1}^{\infty}
    \lambda_i(f) \, \boldsymbol{\Psi}_i(f)\,\boldsymbol{\Psi}_i^*(f) \,
    e^{i 2\pi f \tau} \, df.
\end{equation}
Similarly, the Fourier transform of the fluctuation field can be expanded in SPOD modes:
\begin{equation}
    \hat{\boldsymbol{x}}(f) = \sum_{i=1}^\infty a_i(f) \, \boldsymbol{\Psi}_i(f),
\end{equation}
where $\hat{\boldsymbol{x}}(f)$ is the temporal Fourier transform of $\tilde{\boldsymbol{x}}(t)$, and
$a_i(f) = \boldsymbol{\Psi}_i^*(f)^\top \, \hat{\boldsymbol{x}}(f)$
is the complex-valued modal coefficient.

In this work, SPOD is applied to the inferred high-frequency flow fields to extract energetically dominant, frequency-resolved coherent structures.

\bibliography{aaai2026}

\end{document}